\documentclass[10pt,journal,twocolumn]{IEEEtran}

\usepackage{epsfig}
\usepackage{graphicx}
\usepackage{amsmath,amssymb}
\usepackage{mathsymb}
\usepackage{cite,cs}
\usepackage{bbm}
\usepackage{color,xcolor}
\usepackage{mathtools}
\usepackage{amsfonts}
\usepackage{mathtools}
\usepackage{hyperref}

\hypersetup{
colorlinks=true,linkcolor=blue,filecolor=magenta,
urlcolor=red
}

\usepackage{multirow}
\usepackage{pbox}
\usepackage{wrapfig}
\usepackage{caption}
\usepackage[linesnumbered,algoruled,boxed,lined]{algorithm2e}

\newcommand{\eg}{\emph{e.g., }}

\ifCLASSINFOpdf
\else
\fi

\begin{document}

\title{T-Person-GAN: Text-to-Person Image Generation with Identity-Consistency and Manifold Mix-Up}


\author{Deyin Liu \IEEEauthorrefmark{1}, Lin Wu \IEEEauthorrefmark{2} \IEEEmembership{Senior Member,~IEEE}, Bo Li \IEEEauthorrefmark{3}, Zongyuan Ge \IEEEauthorrefmark{4} 
\IEEEcompsocitemizethanks{\IEEEcompsocthanksitem D. Liu is with Anhui University, China. 
\IEEEcompsocthanksitem L. Wu is with Hefei University of Technology, China.
\IEEEcompsocthanksitem B. Li is with Northwestern Polytechnical University, Xi'an, China.
\IEEEcompsocthanksitem Z. Ge is with Monash University, VIC, Australia.
}
}



\IEEEtitleabstractindextext{
\begin{abstract}
In this paper, we present an end-to-end approach to generate high-resolution person images conditioned on texts only. State-of-the-art text-to-image generation models are mainly designed for center-object generation, e.g., flowers and birds. Unlike center-placed objects with similar shapes and orientation, person image generation is a more challenging task, for which we observe the followings: \textbf{1)} the generated images for the same person exhibit visual details with identity-consistency, e.g., identity-related textures/clothes/shoes across the images, and \textbf{2)} those images should be \textit{discriminant} for being robust against the inter-person variations caused by visual ambiguities. To address the above challenges, we develop an effective generative model to produce person images with two novel mechanisms. Our first mechanism (T-Person-GAN-ID) is to integrate the one-stream generator with an identity-preserving network such that the representations of generated data are regularized in their feature space to ensure the identity-consistency. The second mechanism (T-Person-GAN-ID-MM) is based on the manifold mix-up to produce mixed images via the linear interpolation across generated images from different manifold identities, and we further enforce such interpolated images to be linearly classified in the feature space. This amounts to learning a linear classification boundary that can perfectly separate images from two identities. To address the unstable training issue, we impose an input regularization over the discriminator so as to prevent the training of generator from saturation in the early steps. Our proposed method is empirically validated to achieve a remarkable improvement in text-to-person image generation. Codes are available on \url{https://github.com/linwu-github/Person-Image-Generation.git}
\end{abstract}

}

\maketitle

\IEEEdisplaynontitleabstractindextext

%
\IEEEpeerreviewmaketitle

\section{Introduction}
%
%
%
%


\IEEEPARstart{G}enerative adversarial networks (GANs) \cite{GAN} have shown promising performance in generating sharper images. Following that, text-to-image generation \cite{GAN-CLS,GAWWN} is greatly advanced by a variant of GANs, i.e., the conditional GANs, where one adopts the GANs to generate an image conditioned on the embedding of textual description. Building on such idea, a series of approaches \cite{TAGAN,StackGAN,HDGAN,AttnGAN,StackGAN-pp,TediGAN} have been successfully developed to generate realistic images while subject to the text input with the hypothesis that each text description describes the objects under one specific category.  Among them, the most-widely representative art is StackGAN++ \cite{StackGAN-pp}. Suffering from the bottleneck of GANs to directly generate the high-resolution images due to the bounded capability of gaussian random noise,  which hence inspires turning to stacking multiple generators and discriminators to produce multi-scale images with increasing resolutions. In particular, a bottom-up strategy comes up by initially achieving a coarse layout of the image with lower resolution, where the feature representation is subsequently obtained to combine with embedding of input text, so as to augment the fine-grained details of the generated image in the next round with the higher resolutions.
This paradigm has led to impressive results on simple, well-structured datasets containing varied specific classes of objects (e.g., birds and flowers) cluttered at the image center, known as center-object generation.

\begin{figure*}[t]
\centering
\includegraphics[width=14cm,height=13cm]{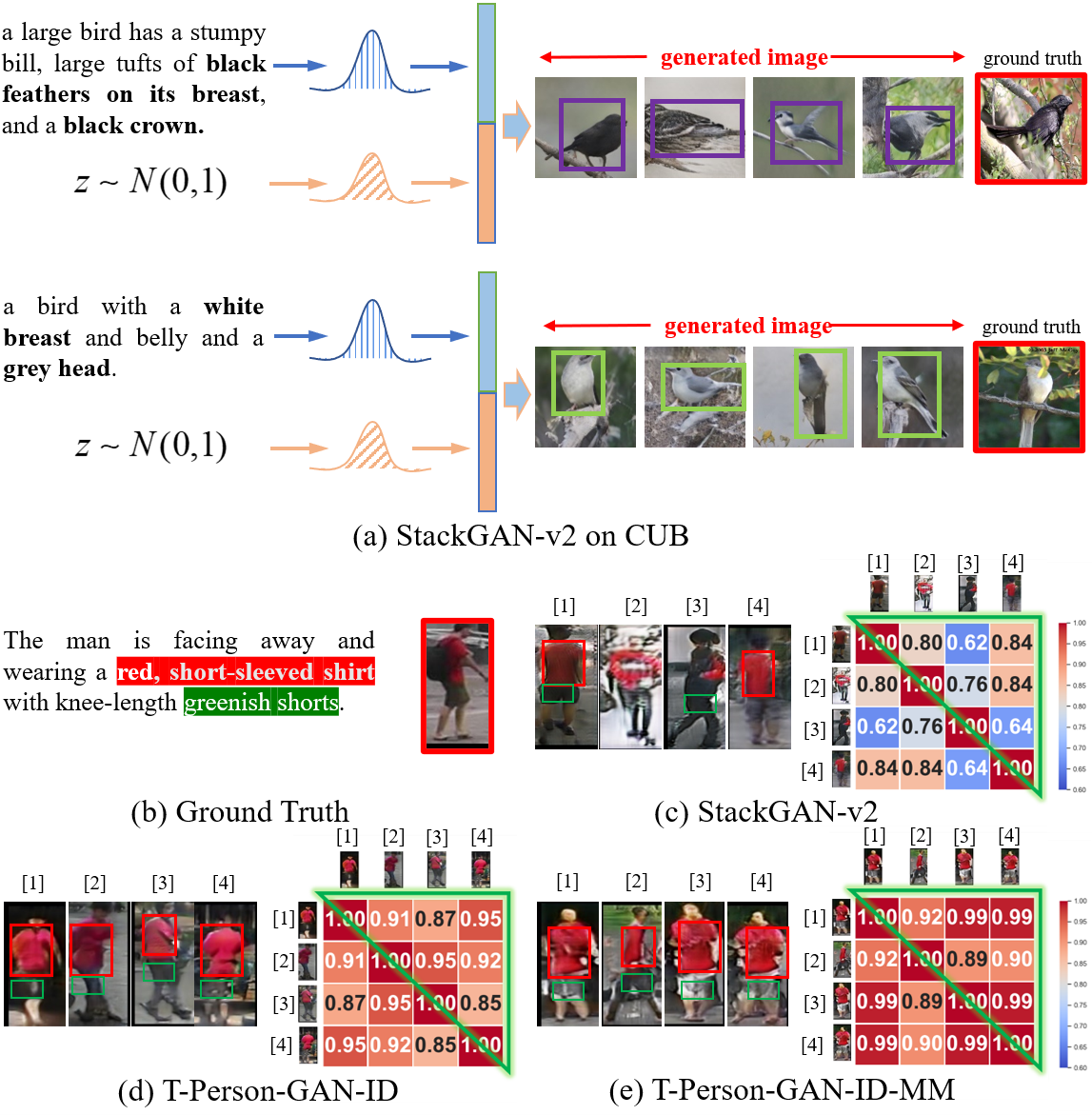}
\caption{ The state-of-the-art center-based object generation method, i.e., Stack-GAN-v2 \cite{StackGAN-pp} mainly increases the randomness of the input signal to generate different birds (a). However, if one adopts randomness in person image generation, it leads to inconsistent regions regarding an identity. As shown in (c), the affinity matrix computed from the generated data in the feature space shows that these images are less correlated. The proposed T-Person-GAN-ID (d) can generate identity-consistent person images, and the proposed T-Person-GAN-ID-MM (e) can further generate discriminant person images against other identities. }\label{fig:center-obj-person}
\end{figure*}

Due to the similar orientations and shapes of center-based objects, a natural question is how to generate diversifying visual objects. To this end, instead of breaking down the bounded randomness of input gaussian noise, the efforts are spent on increasing the randomness of input text signals. Typical example is illustrated in Fig. \ref{fig:center-obj-person} (a), where the most advanced center-object generation method, i.e., StackGAN-v2 \cite{StackGAN-pp}, increases the randomness of the input text via a conditioning augmentation such that more diverse visual objects can be generated. However, such a principle violates in person image generation, which may inadvertently generate inconsistent regions/body parts across the images (see Fig. \ref{fig:center-obj-person} (c)) based on ground truth Fig. \ref{fig:center-obj-person} (b). Unlike the simple center-objects, person image generation requires person images to exhibit consistent regions/body parts such that they are \textit{identity-consistency}. To this end, our proposed method, named T-Person-GAN-ID (Fig. \ref{fig:center-obj-person} (d)), ensures the identity related features, such as colored clothes/shoes, textures and style, to be consistently generated. Thus, the generated data are highly correlated in their feature space, and those generated person images exhibit high identity realism. Also, the generated person images should be \textit{discriminant} for being robust against the inter-person variations caused by visual ambiguities. Henceforth, to synthesize realistic person images, one needs to learn its identity related features during the generation \cite{DG-Net,Joint-GCL,IS-GAN}, and also reinforce the generated images discriminant against other identities.

Generating realistic person images from given texts is of great importance to many downstream tasks such as person appearance transferal and virtual reality. However, this problem is still understudied, we observe the following major facts:  First, unlike the center-object images, the generation should be regularized by an identity-preserving strategy to ensure the generated images to be identity-consistency. However, in inference it is inaccessible to informative identity related features regarding an unknown identity. So a natural question arises: how can we regulate the generator to produce person images with identity-consistency? Inspired by a recent study \cite{HDGAN} that can leverage hierarchical representations from multi-scale representations, we propose to leverage the features of generated data to be correlated in their feature space at identity level. Second, it is challenging to ensure the generated person images discriminant to different persons, i.e., the generative model should achieve the inter-person variations.

\subsection{Our Approach}
In light of the above, our approach expects that generated person images have high identity-consistency; and discriminant against the inter-person variations. To this end, we propose an approach composing two novel mechanisms (namely T-Person-GAN-ID and T-Person-GAN-ID-MM).

Before shedding light on the two mechanisms, like the center-object images, we remark that directly generating a person image with a high resolution, e.g., $256\times 128$, is also difficult due to the brittle training of GANs. This is partially due to the disjoint supports of the data distribution and the model distribution. This problem is more salient when training the GANs to generate higher resolution image from a high-dimensional space. Following the heuristics in center-based object generation \cite{StackGAN-pp,AttnGAN,HDGAN}, we adopt an one-stream architecture to progressively generate multi-scale person images from $64\times 32$, to $128\times 64$, and eventually $256\times 128$. Such one-stream pipeline consists of one generator producing resolution-increasing images, paired with a group of hierarchical discriminators, where each discriminator plays the adversarial games on the corresponding resolution. Our first proposed mechanism, i.e., T-Person-GAN-ID (see Section \ref{ssec:ID-network}) is to effectively incorporate the identity related features, such as colored clothes/shoes, textures and style \cite{DG-Net,Joint-GCL,IS-GAN} , into the generator. Specifically, we make up a set of classifiers on top of the generated images to flatten the class-specific representations, and thus reduce the number of directions with significant variance. This acts as a regularizer to the feature space for the generated images so as to encourage the identity consistency among them.


Different from the center-object images, person images should  be not only identity-consistency within each identity but also discriminant against the inter-person variations caused by the visual ambiguities. Recalling our first mechanism generates the person images that reside on the same identity manifold in the embedding space, we observe that some images are staying nearby the boundary of two identity manifolds due to the visual ambiguities such as illuminations and viewpoint changes, which may result into manifold ambiguity. To address this issue, we propose the second mechanism, i.e., T-Person-GAN-ID-MM (expanded in Section \ref{ssec:manifold-mix-up}) to linearly interpolate the synthesized images from two manifolds, so that the resultant mix-upped data can be \textit{linearly} classified in the embedding space, which essentially amounts to learning linear classification boundaries that can well separate images for different identities, such exemplar is illustrated in Fig. \ref{fig:correlation-t-SNE}. As a byproduct,
it can achieve the compactness of the generated images on the same manifold. To fulfill the goal, a simple \textit{manifold mix-up} strategy comes up to interpolate the generated person images from two identity manifolds at multiple levels (multi-resolutions). Such mix-upped samples are jointly optimized with the generator such that the generated images residing on the manifolds are linearly separable, so that the linear classification on the mix-upped samples are feasible, which further endows the generator with the capacity of synthesizing more discriminant images for different identities. To properly train these mix-upped samples, we develop a teacher-student type supervision with dynamic soft labeling.

In summary, our generated model comprising two mechanisms can well preserve the intra-person identity consistency and inter-person variations. Finally, to stabilize multiple discriminators into one-stream generator, we add a regularization term into the discriminator objective (see Section \ref{ssec:stable-training}), to control the Lipschitz constant of the discriminator over the input samples, to facilitate the generator with more stable gradients.
\textit{Orthogonal to conventional StackGAN-v2 aimed at center-object image generations, our developed pipeline focuses on the person images generation given text input, which, together with StackGAN-v2, belong to and enrich the GANs paradigm for image generations.}

The contributions of this paper are summarised below: \textbf{1)} We propose an end-to-end approach to generate highly realistic person images from texts only, which mainly consists of two major mechanisms, i.e., identity-preserving and manifold mix-up criterion.  \textbf{2)} The identity-preserving mechanism exploits the feature space of generated data w.r. t identity consistency, and the interpolation mechanism via manifold mix-up is further proposed to generate more robust person images against visual ambiguities. \textbf{3)} Extensive experiments are conducted to demonstrate the superiority of our method in text-to-person image generation.

The rest of the paper is structured as follows. Section \ref{sec:related} describes related works. Preliminaries are presented in Section \ref{sec:preliminary}. Section \ref{sec:method} details the problem setup and the proposed method with training procedure. Section \ref{sec:experiment} reports both qualitative and quantitative experiments, and Section \ref{sec:con} concludes this paper.

\section{Related Work}\label{sec:related}

In this section, we review literature on topics of generative adversarial networks, textual-to-image synthesis, person image generation and sample mix-up.

\subsection{Deep Generative Models}

Generative adversarial networks (GANs) \cite{GAN} were originally presented as a means of learning a generative model which captures an arbitrary data distribution from a particular domain. The input to the generator $G$ is a noise vector $\textbf{z}$ drawn from a latent distribution, such as a multivariate Gaussian. With the resurgence of GANs, several works have extended GANs in different aspects. For instance, DCGAN \cite{DCGAN} extends GANs by leveraging deep neural networks and provides the best practices for training GANs. InfoGAN \cite{InfoGAN} extends GANs by additionally maximizing the mutual information between interpretable latent variables and the generator's distribution. More recent studies extend GANs by feeding auxiliary information, such as class labels \cite{CGAN}, sentence descriptions \cite{GAWWN,GAN-CLS,StackGAN} into both the generator and discriminator. Meanwhile, the theory of GANs is investigated and the findings show that the Jenson-Shannon divergence optimized by the standard GANs leads to instability and mode collapse \cite{Mode-seeking}. To combat the unstable training of GANs, Wasserstein-GAN (WGAN) \cite{WGAN} is proposed to optimize an efficient approximation of the Wasserstein distance. While WGAN attains better theoretical property than the vanilla GAN, it still suffers from the gradient exploding and vanishing problem because it uses weight clipping to enforce the Lipschitz constraint on the discriminator. Although those papers have demonstrated realistic images, they are short in generating discriminative features suitable for classification purpose. Also, those generative models are limited in comparing the probability distribution between real data and the generated data.

\subsection{Text-to-Image Generation}
Text-to-image synthesis emerges as an interesting application of GAN. Reed \etal \cite{GAN-CLS} introduce a method to generate $64^2$ images by using an image-text matching aware adversarial training. To improve fine-grained details, several methods are developed to address the association between words and output regions. They are mainly driven to determine where and what the content should be generated \cite{GAWWN,AttnGAN,StackGAN,SD-GAN} by using, e.g., attention-driven or hierarchically cascaded networks. For instance, Xu \etal \cite{AttnGAN} propose to deploy the attention mechanism on each generator to determine relevant words and the generated object parts. One fundamental difference of our model from those text-to-image methods \cite{StackGAN,HDGAN} is that they typically incorporate multiple generators into the adversarial cycle in order to produce the samples with fine-grained details. In the way of using multiple generators, the balance of training between the generators and its corresponding texts is not easily guaranteed. In contrast, we propose an one-stream generative architecture with multiple discriminators as hierarchically nested objective. More recently, a variant of GANS, i.e., conditional GANs have made great progress in learning a continuous textual embedding from a low-dimensional manifold to a complex real image distribution \cite{StackGAN-pp,HDGAN}. However, all above approaches are developed based on center-object generation, which is not applicable to person image generation. Some other recent approaches such as SSA-GAN \cite{SSA-GAN}, RiFeGAN \cite{RiFeGAN} and MirrorGAN \cite{MirrorGAN} use different priors, e.g., masks and enriched textual knowledge to improve the generation realism. In this paper, we aim to generate person images conditioned on texts only.

\subsection{Person Image Generation}

Current generative models for human images are often conditioned on a source image and a target pose specification. This pipeline have been investigated with great attention \cite{ma2017pose,Conditional-DGPose,ClothNet-Body,Deformable-GANs,PN-GAN,Variational-U-Net}. One typical approach is \cite{ClothNet-Body} (ClothNet-Body), which presents a generative model based on CVAEs for clothes of segmented people conditioned on the pose. Their generative model requires an image-to-image translation network \cite{Img-Img-Translation} to render natural images. Ma \etal \cite{ma2017pose} deploy the U-Net based architecture \cite{Fusion-net} to allow to synthesize person images with any arbitrary pose. The input of their model is a conditioning image of the person and a target new pose defined by 18 joint locations. However both \cite{ClothNet-Body} and \cite{ma2017pose} work in two stages: pose generation and texture refinement, due to the challenges of training a complete end-to-end framework to handle poses and human appearance simultaneously. To overcome this challenge, a large body of research \cite{Conditional-DGPose,Deformable-GANs,PN-GAN} propose generative models by disentangling pose from other latent factors of variations, e.g., background and clothing. These approaches show that a single-stage end-to-end training framework can be developed for obtaining higher qualitative results. Different from aforementioned works, we are motivated to consider a more challenging setting: generating human images only conditioned on natural language descriptions. This is to eliminate the training dependency on a reference image from a particular subject, which is often difficult to be obtained in real applications. Also, without being limited to human poses, the image generation process can produce a lot of images with full descriptions on each identity. To this end, we propose to use continuous variables from texts to generate human images on manifolds. And the discriminative power of generation is achieved by leveraging the perceptual loss of CNN features into the GAN loss. \cite{CLSWGAN} also proposes a WGAN based formulation that uses a discriminative supervised loss function, in addition to the unsupervised adversarial loss. In this model, the supervised loss enforces the WGAN generator to produce samples that are correctly classified according to a pre-trained classifier of seen classes.


\section{Preliminaries}\label{sec:preliminary}

Our model follows the principle of Generative Adversarial Networks (GANs) \cite{GAN}, comprising a generator $G$ and a discriminator $D$ that compete in a two-player min-max game. The discriminator $D$ is optimized to distinguish synthesized images from real images, meanwhile $G$ is trained to fool $D$ by synthesizing fake images. Overall, the optimal $G$ and $D$ can be obtained throughout the following two-player min-max game,
\begin{equation}
    G, D =\arg\min_G \max_D V(G, D, I, \mathbf{z}),
\end{equation}
where $I$ and $\textbf{z}\sim \mathcal{N} (0,1)$ denote the training images and random noises, respectively. $V(\cdot)$ is the overall GAN objective, which usually takes the form of $\mathbb{E}_{I\sim p_{data}} [\log D(I)]+\mathbb{E}_{\mathbf{z}\sim p_z} [\log (1-D(G(\mathbf{z})))]$.

Conditional GANs are variants of GANs, where both the generator and discriminator receive additional conditioning variables $\textbf{c}$ (\eg $\textbf{c}$ can indicate a semantic label, a source image, or textual descriptions), yielding $G(\textbf{z},\textbf{c})$ and $D(I, \textbf{c})$. This formulation allows $G$ to generate images conditioned on $\textbf{c}$. Our network is developed based on conditional GANs by including the text into $G$ to generate a person image with growing resolutions.

\section{Our Method}\label{sec:method}
We propose a generative adversarial network to generate person images from \textit{texts only}, which works based on one-stream generative model, including 1) one generator to produce increasing-resolution images, and 2) a group of hierarchical discriminators at multi-scale intermediate layers to play multiple adversarial games, i.e., differentiate real/fake patches as well as real/fake image-text pairs. Note that this unification of generator and discriminator is different from the conventional multi-stage framework \eg StackGAN++ \cite{StackGAN-pp}, by stacking multiple generators and discriminators for optimizing different resolution distributions. To simultaneously achieve intra-person identity consistency and inter-person variations, we propose two mechanisms: 1) an identity-consistency network over intermediate layers of the generator to ensure the identity-consistent (namely T-Person-GAN-ID); and 2) a linear interpolation amid synthetic images via manifold mix-up to ensure the discriminant against inter-person variations (namely T-Person-GAN-ID-MM).

\subsection{One-Stream Generator with Multi-Adversarial Objectives}\label{ssec:one-stream}

\begin{figure*}[hbt]
\centering
\includegraphics[width=1.5\columnwidth]{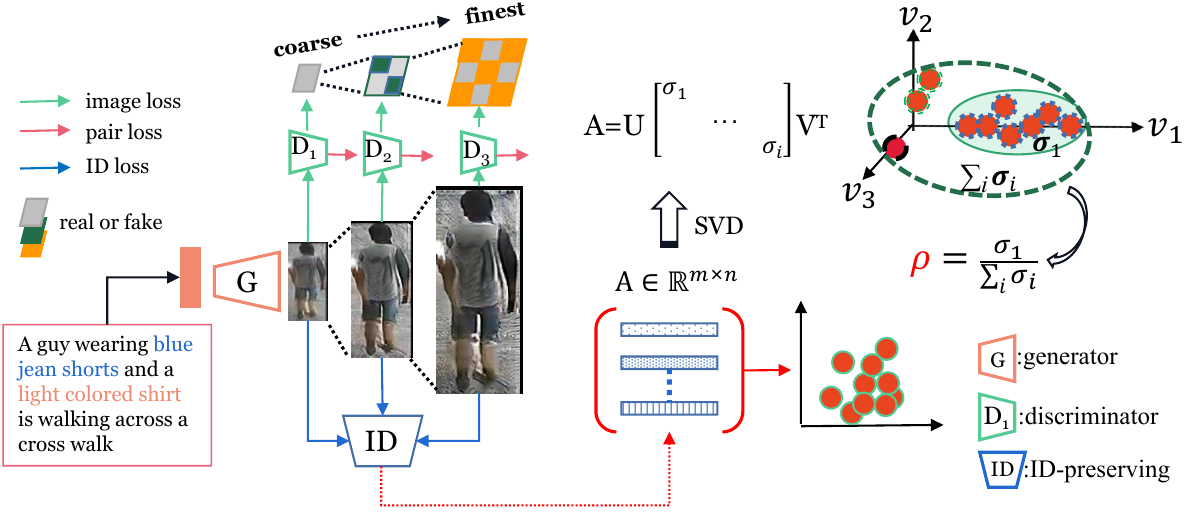}
\caption{The proposed person image generation from texts. The network is composed of an one-stream generator, hierarchical discriminators and an identity-preserving network. For each intermediate output, the corresponding discriminator $D_i$ computes the matching-aware pair loss with given texts as well as the local image loss from coarse to finest. The identity-preserving technique regularizes the feature space of mid-level representations of generated images. Such a regularization is shown to have strong correlation, i.e., the correlation ratio is defined as $\rho=\frac{\sigma_1}{\sum_i \sigma_i}$, where $\sigma_i$ denotes the $i$-th largest singular value of the generated data vector matrix. Best viewed in color.}\label{fig:one-stream-generator}
\end{figure*}

We adopt the one-stream generator $G$ to produce images at increasing resolutions, where a group of discriminators at multi-scale intermediate outputs can play multiple adversarial games. Specifically, $G$ takes texts as its input and produces multiple intermediate outputs:
\begin{equation}
    X_1, \ldots, X_s = G(\mathbf{c}, \mathbf{z}),
\end{equation}
where $\mathbf{c}\sim p_{data}$ denotes a sentence embedding (e.g., it can be generated from a pre-trained char-RNN text encoder \cite{GAN-CLS}). Instead of relying on a fixed conditioning variable $\mathbf{c}$, we follow the practice of StackGAN++ \cite{StackGAN-pp} and sample a stochastic vector for $\mathbf{c}$ from an independent Gaussian distribution $\mathcal{N}(\mu(\phi_t), \sum(\phi_t))$, where $\phi_t$ is the text embedding of a text description $t$. The mean $\mu(\phi_t)$ and diagonal covariance matrix $\sum(\phi_t)$ are functions of the text embedding $\phi_t$. The set of $\{X_1,\ldots, X_s\}$ are images with growing resolutions, where $X_s$ represents the final output with the highest resolution. In our case, we produce person images with increasing resolutions from $64 \times 32$ to $128\times 64$, and ultimately $256\times 128$.

Being conditioned on \textbf{c}, we train $G$ with hierarchical discriminators $D_i$, ($i=1,\ldots,s$), at different scales $s$ by minimizing both $\mathcal{L}_D$ for $D$, and $\mathcal{L}_G$ for $G$ via the following objectives:
\begin{equation}\label{eq:min-max}
\begin{split}
& \mathcal{L}_D= \sum_{i=1}^s \underbrace{\mathbb{E}[D_i(I_i)-\mathbb{I}] +\mathbb{E} [ D_i(X_i)]}_\text{image loss}  \\
& + \underbrace{\mathbb{E}[D_i(I_i, \textbf{c}_i]-\mathbb{I}) +\mathbb{E} [ D_i(X_i, \textbf{c}_i)] + \mathbb{E} [D_i (I_i, \bar{\textbf{c}}_i)]}_\text{pair loss}\\
& \mathcal{L}_G= \sum_{i=1}^s \mathbb{E} [ D_i (G(\textbf{c}, \mathbf{z})_i)-\mathbb{I}] + \mathbb{E} [ D_i(G(\mathbf{c}, \mathbf{z})_i, \mathbf{c}_i) - \mathbb{I}],
\end{split}
\end{equation}
where $I_i$ indicates a real image at different scales. $\mathbb{E}[(x-\mathbb{I})^2]$ is the mean-square loss, where the shape of $x$ and $\mathbb{I}$ varies w.r.t the output size of the generator. Hence, minimizing $\mathcal{L}_D$ involves two loss branches: the image loss, i.e., $D_i(X_i)$ and $D_i(I_i)-\mathbb{I}$, together with the matching-aware pair loss, i.e., $D_i(I_i, \textbf{c}_i)$, $D_i(X_i, \textbf{c}_i)$ and $D_i (I_i, \bar{\textbf{c}}_i)$. During training, each $D_i$ takes real images and their corresponding texts as positive sample pairs ($\{I_i, \textbf{c}_i\}$), whereas negative pairs consist of two groups: the real images with mismatched text embeddings $\{I_i, \bar{\textbf{c}}_i\}$ and the synthetic images with their corresponding text embeddings $\{X_i, \textbf{c}_i\}$. The output of matching-aware pair loss is a real value. For the image loss, following \cite{HDGAN}, we adopt the hierarchical global-to-local structure to compute the loss at different resolutions (see Fig. \ref{fig:one-stream-generator}). As suggested in \cite{HDGAN,Img-Img-Translation}, low-to-high resolution discriminators at hierarchy are supposed to focus on global structures and local image details, respectively. Specifically, for each $D_i$, we compute the 2-dimensional probability map $O_i \in \mathbb{R}^{R_i\times R_i}$, where the shape of $O_i$ varies in accordance to the resolution. For example, $R_i=1$ refers to the global range. The local loss $R_i$ can be adjusted to control the receptive field of each element in $O_i$, which determines whether a corresponding local patch is real or fake.

To optimize $G$, Eq. \eqref{eq:min-max} minimizes $\mathcal{L}_G$ by involving two terms: $D_i(G(\textbf{c}, \mathbf{z})_i)$ and $D_i(G(\mathbf{c}, \mathbf{z})_i, \mathbf{c}_i)$, which try to fool the discriminator with a fake image, and a pair of fake image with its corresponding text. As suggested in \cite{StackGAN-pp}, to enforce the smooth sampling over the text embedding distribution, a regularization term, $D_{KL}(\mathcal{N}(\mu(\phi_t), \sum(\phi_t))|| \mathcal{N}(0, \mathbf{I}))$, \textit{i.e.,}  a Kullback-Leibler divergence between the standard Gaussian distribution and the conditioning Gaussian distribution) should be added into the objective of the generator during training.

\subsubsection{Model Architecture}

Following \cite{HDGAN}, $G$ can be simply seen as a CNN backbone, which is composed of three modules, namely $K$-repeated residual blocks, stretching layers, and linear compression layers. Analogous to the standard residual block \cite{Resnet}, each res-block from the $K$-repeated res-blocks contains two convolutional layers with batch normalization (BN) \cite{BN} and ReLU. To reduce the sparse gradients, we remove the ReLU after the skip-addition of each residual block. The stretching layer, which contains a scale-2 nearest up-sampling layer followed by a convolutional layer with BN+ReLU, is to change the size and dimension of feature maps. The linear compression layer is a single convolutional layer followed by a Tanh function to directly compress feature maps to the RGB space. To compute the text embedding $\phi_t$, we adopt the BERT model \cite{BERT} to produce a 768-dim text vector, which is further combined with the Conditioning Augmentation (CA) \cite{StackGAN-pp} to yield a final 128-dim embedding. The generator starts from a $128\times 4\times 4$ embedding anchored on $B$ $K$-repeated res-blocks, which are connected by $B-1$ in-between stretching layers until the resolution of the feature maps equal to the targeting resolution. For example, for a $256\times 128$ resolution, we set $K=1$ and apply $B=6$ 1-repeated res-blocks, together with 5 stretching layers.

In our model, the discriminator contains consecutive convolutional layers with stride-2 and BN+LeakyReLU. On top of the discriminator, two branches are made:  a fully convolutional layer to produce the probability map $O_i$ at each resolution, and classify each location as real or fake, with another branch that concatenates a $512\times 4 \times 4$ and a $128\times 4\times 4$ embedding (formed by replicating the 128-dim text embedding), and subsequently uses an $1\times 1$ convolution to fuse the text and image features, followed by a $4\times 4$ convolution to classify an image-text pair to be real or fake.

\subsection{Mechanism 1: Identity-Consistency Generation}\label{ssec:ID-network}

We argue that the generated person images for the same identity at different resolutions should exhibit identity-consistency, i.e., always present the identity-related details, such as clothing/shoes color, texture and style. To this end, we deploy an one-stream architecture coupled with multiple discriminators. The discriminators are hierarchically nested into the generator for resolution-increasing images. Such an architecture facilitates the consistency by incorporating identity-related features into the multi-scale outputs. Fig. \ref{fig:one-stream-generator} shows our one-stream generative model, which integrates the hierarchical discriminators and an \textit{identity-consistency} network. To embed the identity-consistency into the multi-scale outputs, we add an identity regularization into the generative objective, which is formulated as $\mathcal{L}_G$ below:
\begin{equation}\label{eq:generator-obj}
    \begin{split}
& \mathcal{L}_G= \sum_{i=1}^s \mathbb{E} [-\log D_i (G(\textbf{c}, \mathbf{z})_i)] + \mathbb{E} [ D_i(G(\mathbf{c}, \mathbf{z})_i, \mathbf{c}_i)]\\
&+ \lambda_1 \mathcal{L}_{CE}(G(\textbf{c}, \mathbf{z})_i, y),
    \end{split}
\end{equation}
where $\mathcal{L}_{CE}(G(\textbf{c}, \mathbf{z})_i, y)$ denotes the cross-entropy loss applied on the generated image at the $i$th scale , and $\lambda_1$ is the balance parameter. Specifically, $\mathcal{L}_{CE}(G(\textbf{c}, \mathbf{z})_i, y)$ is computed as follows:
\begin{equation}
   \mathcal{L}_{CE}(G(\textbf{c}, \mathbf{z})_i, y) =-\frac{1}{C}\sum_{k=1}^C y_k \cdot \log (\hat{y}_k),
\end{equation}
where $C$ denotes the total number of identities; $y_k$ is the ground truth label and $\hat{y}_k$ is the predicted label severing as the output from a soft-max classifier over an input $G(\textbf{c}, \mathbf{z})_i$ represents the $i$th scale. Intuitively, the cross-entropy loss aims to minimize the mis-classification of the generated images w.r.t the class of identities. We remark that explicitly minimizing $\mathcal{L}_{CE}(G(\textbf{c}, \mathbf{z}),y)$ is equaling to encouraging $G$ to produce person images that possess identity-related features. As such, the soft-max classification $\hat{y}_k$ can make the correct prediction on each identity $k$. Intuitively, forwarding the generated images into the identity classification module (parameterized by $\mathcal{L}_{CE}(\cdot)$) can ensure these images to be identity-correlated. Meanwhile, by jointly optimizing with $\mathcal{L}_{CE}(\cdot)$, the generator can receive the identity-related gradients to generate different person images under the same identity with high identity-consistency.

\begin{figure}[t]
\includegraphics[width=1\columnwidth]{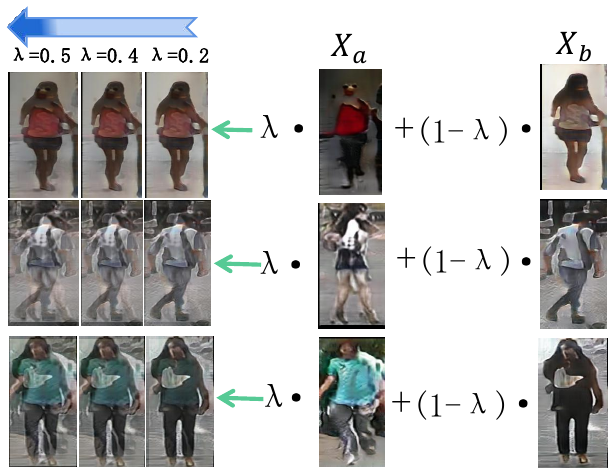}
\caption{The mix-upped image by linear interpolation between generated images from two identities. For a generated image $X_b$ from identity $b$, the mix-upping outcome tends to show more patterns from identity $a$, as the modulation parameter $\lambda$ increases. Best viewed in color.}\label{fig:mix-up-showcase}
\end{figure}

\subsubsection{High Correlation between Synthesized Images}

Recalling section \ref{ssec:ID-network}, where the rationale of the identity-preserving network is to encourage the identity consistency across person images for the same identity at different resolutions. To do this, the term $\mathcal{L}_{CE}$ in Eq.\eqref{eq:generator-obj} comes up to regularize the intermediate outputs of the generator in the feature space. Herein \textit{one may wonder how} could we evaluate the consistency of these generated images, and can we further explicitly quantify it? To answer this question, we perform the singular value decomposition (SVD) on the data matrix formed by the feature vectors of generated images extracted from the ID network, as shown in Fig. \ref{fig:one-stream-generator}.

Mathematically, suppose $A\in \mathbb{R}^{m\times n}$ to be the data matrix with $m$, $n$ dimensional image vectors regarding one person identity, we decompose $A$ as $A=U \Sigma V^T$, where $U$ and $V$ are the orthonormal vectors and $\Sigma$ is a diagonal matrix (the $i$th entry of $\Sigma$, denoted as $\sigma_i$,  represents the $i$th singular value). The hypothesis of taking the $k$ largest singular values of $A$ (replacing the rest with zeros in $\Sigma$) and recomputing $U\Sigma V^T$ implies the provably-best $k$-rank approximation to the matrix. Based on that, the summation of the first $k$ singular values normalized by all the singular values, i.e., $\frac{\sum_i^k \sigma_i}{\sum_i \sigma_i}$ reveals \textit{how much information those singular values contain}. For example, if $k$=1, then the ratio $\rho=\frac{\sigma_1}{\sum_i \sigma_i}$ can measure the correlation between those samples: the higher $\rho$ is, the stronger the correlation is, further lead to the higher rank-1 approximation to $A$. We empirically find that $\rho$ is greater than 0.25 for all identities for Mechanism 1. (see more details in section \ref{sec:experiment}).

Up-to-now, we have achieved the identity-consistency across multi-scale person images from the same identity (i.e., intra-person closeness). As aforementioned, person image generation is fundamentally different from center-object generation in the sense of requiring generated images discriminant against inter-person variations caused by visual ambiguities. To this end, we propose a manifold mix-up mechanism to achieve the generated images discriminant against inter-person variations in the next section. 

\subsection{Mechanism 2: Manifold Mix-Up for Discriminant Person Image Generation}\label{ssec:manifold-mix-up}

The first mechanism can produce identity-consistency person images, which, as we observed, reside on the identity manifolds. However, due to the visual ambiguities, some images may stay nearby the boundary of two manifolds, causing manifold ambiguities. To this end, we propose to linearly separate these manifolds by learning the linear boundary between them. For the first glance, one may combine the above identity-consistency mechanism with a contrastive loss \cite{SD-GAN} to achieve both the intra-person compactness and the inter-person difference. However, such a strategy has to measure the distance between the generated data with potential cross-identities during generation; worse still, this loss cannot induce the well-clustered images in these manifolds so as unable to separate the images via the linear fashion.

\begin{figure}[t]
\includegraphics[width=1\columnwidth]{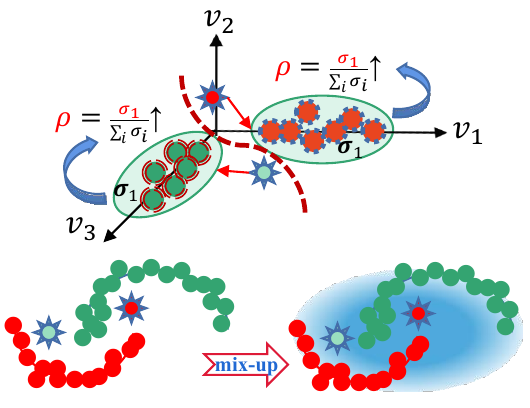}
\caption{Mix-up samples in their embedding space. The mix-up formulation allows more stronger correlation between samples (the ratio $\rho$ is higher), and also robustness between different classes (discriminant). Best viewed in color.}\label{fig:mix-up-embedding}
\end{figure}

To address the challenge above, we propose to interpolate the generated images between varied identity manifolds, and linearly classify these mix-upped images in the embedding space. By doing so, we learn linear boundary between manifolds so as to well differentiate images for different identities (This can demonstrated by the t-SNE visualization in Fig. \ref{fig:correlation-t-SNE}). More specifically, for a generated image $X_a^i$ over the $i$th scale from identity $a$, and another generated image $X_b^i$ from identity $b$ ($a\neq b$), we apply the manifold mix-up formulation to produce the virtual image $\bar{X}^i$ as follows:
\begin{equation}\label{eq:mixup}
\bar{X}^i= \lambda \cdot X_a^i + (1-\lambda) \cdot X_b^i,
\end{equation}
where $\lambda \sim Beta(\alpha,\alpha)$ for $\alpha \in (0, \inf)$, and $\lambda \in [0,1]$. The mix-up parameter $\alpha$ controls the strength of interpolation between the synthetic-synthetic pairs. $\alpha$ is empirically set to 0.5. Intuitively, Eq. \eqref{eq:mixup} aims to leverage these interpolations across identities, which is modulated by $\lambda$. We offer an illustration of manifold mix-up on generated images in Fig.\ref{fig:mix-up-showcase}.

To use the combinations of feature representations of generated data, we also conduct the same linear interpolation between the pair of labels, leading to the mixed data with soft targets. For example, the soft label for the mixed data $\bar{X}^i$ is $\bar{y}= \lambda \cdot a + (1\lambda) \cdot b$. To impose the mix-up loss into the network, we add the following additional regularization to the generator optimization:
\begin{equation}\label{eq:regularize-D}
R(G)=\sum_{i=1}^s \mathcal{L}_{TS}(F (\bar{X}^i), \bar{y}),
\end{equation}
where the $\bar{y}$ serves as a soft-label. $F(\cdot)$ is a soft-max classification, along with the loss function $\mathcal{L}_{TS}(,)$. Hence, the generator objective function becomes:
\begin{equation}
    \begin{split}
& \mathcal{L}_G= \sum_{i=1}^s \mathbb{E} [-\log D_i (G(\textbf{c}, \mathbf{z})_i)] + \mathbb{E} [ D_i(G(\mathbf{c}, \mathbf{z})_i, \mathbf{c}_i)]\\
&+ \lambda_1 \mathcal{L}_{CE}(G(\textbf{c}, \mathbf{z})_i, y) + \lambda_2 R(G),
    \end{split}
\end{equation}
where $\lambda_1$ and $\lambda_2$ are balance parameters for two regularization terms, which are empirically tuned as $\lambda_1=0.5$ and $\lambda_2=0.1$. We argue that such manifold mix-up between generated data can squash the representation variance, leading to reduced directions of sample volume in the feature space. Inspired by this, we jointly optimize the generator and perform the linear classification on these interpolated images, so as to encourage the linear arrangement of generated data representations. As a result, the identity-representations are flattened into a minimal amount of directions of variation. In term of the feature space, the representation computed from the generated images occupy a smaller volume, yielding more stronger correlations within the same identity. Thus, images at different identities are well separated, as shown in Fig. \ref{fig:mix-up-embedding}.

\subsubsection{Training with Mix-Up Samples}
To encourage the generated images to be discriminant against different identities, we optimise the network by training a classifier over mix-upping data at multi-scales. This classification loss can be interpreted as a regularizer encouraging the generator to construct discriminant features. The generative model is designed to leverage synthetic examples to linearly classify the identity manifolds.

Recent studies indicate the possibility to treat the generated images as training samples \cite{MpRL,DG-Net,PN-GAN}. However, due to the inter-class variations in the interpolated images, it is difficult to determine the exact label for these composed images. Thus, we adopt the teacher-student type supervision with the dynamic soft labeling. Specifically, we use a teacher model to dynamically assign a soft label to $\hat{X}_a^b$, depending on its interpolation from $X_a$ and $X_b$, which is further modulated by $\lambda$. The higher $\lambda$ is, the more $\hat{X}_a^b$ similar to $X_a$. The teacher model is a simple CNNs trained identification loss on the original training set. To train the mix-up model for more discriminant images, we minimize the KL divergence between the probability distribution $p(\hat{X}_a^b)$ predicted by the mix-up module (student) and the probability distribution $q(\hat{X}_a^b)$ predicted by the teacher:
\begin{equation}
   \mathcal{L}_{TS}= \mathbb{E}[-\sum_{m=1}^M q(m|\hat{X}_a^b) \log(\frac{p(m|\hat{X}_a^b)}{q(m|\hat{X}_a^b)})],
\end{equation}
where $M$ is the number of identities. In comparison with the fixed one-hot label \cite{PN-GAN}, or static smoothing label \cite{LSRO}, this dynamic soft labeling is valid in our scenario since each synthetic image is formed by interpolating from images with different identities. However, training the above two mechanisms with a GAN is difficult, due to the unstable training from GANs, partially due to the disjoint between the synthetic data distribution and the real data distribution. This problem is especially severe for person image generation, which incurs the instability due to the inaccuracy of discriminator in estimating the density ratio in such high-dimensional space. Moreover, in multi-resolution generation, when the support of the model distribution and that of the target distribution in low-resolution are disjoint, there exists a discriminator that can perfectly distinguish them. Once such discriminator is learned, the training for generator will terminate since the derivatives of such discriminator w.r.t the input will be zeros. To address this challenge, we propose a more stable training strategy for person image generation.

\subsection{More Stable Training for Person-Image Generation}\label{ssec:stable-training}

To prevent the early-stopping of generator in multi-resolution generation, we propose an \textit{input based regularization} that allows for an easy formulation based on the input sample. Recalling that the conventional formulation of GANs takes the form $\min_G \max_D V(G,D)=\mathbb{E}_{x\sim q_{data}} [\log D(x)]+ \mathbb{E}_{x'\sim p_G}[\log(1-D(x'))]$, where $q_{data}$ is the data distribution, and $p_G$ is the model (generator) distribution to be learned through the adversarial min-max optimization. It is known that, for a fixed generator $G$, the optimal discriminator is given by $D^*_G(x)=\frac{q_{data}(x)}{q_{data}(x)+p_G(x)}=\mbox{sigmoid}(f^*(x))$, where $f^*(x)=\log q_{data}(x)-\log p_G(x)$. However, the derivative of $ f^*(x)$, i.e., $\triangledown f^*(x)$ could be unbounded. Hence, it prompts us to impose a regularity condition on $\triangledown f^*(x)$.

Recent study \cite{Spectral-norm} suggests that an effective approach to regularize the derivative is to control the Lipschitz constant of the discriminator by adding a regularization term over the input examples $x$ (or $x'$). We follow this line of research and search for the discriminator $D$ from the set of $K$-Lipschitz continuous functions, that is,
\begin{equation}\label{eq:K-Lipschitz}
    \arg\max_{||f||_{Lip}}\leq K,
\end{equation}
where $||f||_{Lip}$ denotes the smallest value $Q$ such that $\frac{||f(x)-f(x')||}{||x-x'||_2}\leq Q$ for any $x$ and $x'$. As a result, we place the $K$-Lipschitz (we set $K=1$) constant on $D$ by modifying the objective function where the regularizer rewards the function for having a local 1-Lipschitz constant: $||\triangledown f^*(x)||_2=1$. Hence, we have the following objective:
\begin{equation}
\begin{split}
    & \min_G \max_D V(G,D)=\mathbb{E}_{x\sim q_{data}} [\log D(x)]+ \mathbb{E}_{x'\sim p_G}[\log(1-D(x'))] \\
    & +\mathbb{E}_{\hat{x}} [(|| \triangledown_{\hat{x}} D(\hat{x}) ||_2-1)^2],
\end{split}
\end{equation}
where $\hat{x}$ is generated by interpolating $x'$ from $p_G$ and $x$ from $q_{data}$. This direct regularization on $D$ sets the its \textit{derivatives bounded}, and simultaneously addresses the disjoint space between the support of $p_G$ and $q_{data}$. Thus, the proposed input based regularization can stabilize the training of GANs.




\section{Experiments}\label{sec:experiment}


In this section, we perform extensive experiments to evaluate the proposed method both quantitatively and qualitatively. We also conduct ablation studies in the following aspects: what is role of ID features in person image generation, how discriminant the generated images, and how the input normalization on discriminator can stabilize the training. 

\subsection{Dataset}

We perform experiments on the CUHK Person Description Dataset (CUHK-PEDES) \cite{GNA-RNN}. The \textbf{CUHK-PEDES} dataset is the only yet largest benchmark for person search with natural language description, which consists of 40,206 images of 13,003 persons. Each image is annotated with two sentence descriptions, such that a total of 80,412 sentences are collected. Following the data split in \cite{GNA-RNN}, the training set has 34,054 images (68,126 textural descriptions) of 11,003 persons, the validation has 3,078 images (6,158 textural descriptions) of 1,000 persons, and the test set has 3074 images (6,156 textural descriptions) of 1,000 persons. We uniformly resize all images to be $256\times 128$. 


\begin{table*}[t]
\centering
\caption{The comparison results on FID, IS and VS-Similarity on the dataset CUHK-PEDES. For the FID, the lower value $\downarrow$ indicates better performance. For both IS and VS-Similarity, a higher score $\uparrow$ represents a better performance. The best results are in bold font.}\label{tab:main-evaluations}
\begin{tabular}{|rc|c|c|c|}
\hline
\multicolumn{2}{|c|}{Method} & \multicolumn{3}{c|}{CUHK-PEDES} \\
\cline{3-5}
&& \textbf{FID} (Realism)$\downarrow$ & \textbf{IS} (Diversity)$\uparrow$ &\textbf{VS-Similarity} (Visual-semantic consistency)$\uparrow$\\
\cline{3-5}
\hline
Ground Truth &- & 0.20 & 4.54 & 0.27 \\
GAWWN \cite{GAWWN} & NIPS-2016 & 109.69$\pm$87 & 2.86$\pm$.12 & 0.10$\pm$.02\\
GAN-INT-CLS \cite{GAN-CLS} & ICML-2016 & 107.23$\pm$85 & 2.94$\pm$.14 & 0.09$\pm$.03 \\
HDGAN \cite{HDGAN} & CVPR-2018 & 61.62$\pm$.69 & 3.85$\pm$.08 & 0.14$\pm$.06 \\
AttnGAN \cite{AttnGAN} & CVPR-2018 & 76.48$\pm$.56 & 3.68$\pm$.12 & 0.14$\pm$.07 \\
MirrorGAN \cite{MirrorGAN} & CVPR-2019 & 60.96$\pm$.63 & 3.69$\pm$.16& 0.19$\pm$.05\\
StackGAN-v2* \cite{StackGAN-pp} & IEEE-TPAMI-2019 & 68.14$\pm$.73 & 3.53$\pm$.08 & 0.13$\pm$.07\\
RiFeGAN \cite{RiFeGAN} & CVPR-2020 & 56.47$\pm$.61 & 3.72$\pm$.22& 0.19$\pm$.03\\
DF-GAN \cite{DF-GAN} & CVPR-2022 & 56.10$\pm$.52 & 3.79$\pm$.17& 0.18$\pm$.04\\
SSA-GAN \cite{SSA-GAN} & CVPR-2022 & 58.45$\pm$.49 & 3.71$\pm$.17& 0.18$\pm$.04\\
T-Person-GAN-ID & Ours &\textbf{48.04}$\pm$.67 & \textbf{3.94}$\pm$.09 & 0.18$\pm$.03  \\
T-Person-GAN-ID-MM & Ours &\textbf{47.81}$\pm$.09 & \textbf{3.96}$\pm$.14 & \textbf{0.21}$\pm$.03 \\
\hline
\end{tabular}
\end{table*}

\subsection{Evaluation Metrics}

We evaluate our method and competitors on three quantitative metrics: Fr$\acute{e}$chet Inception Distance (FID) \cite{FID}, Inception Score (IS) \cite{Inception-score} and Visual-Semantic Similarity (VS-Sim). (1) \textbf{Fr$\acute{e}$chet Inception Distance}\cite{FID} captures the similarity of generated images to real ones by considering the difference of two Gaussians (synthetic and real-world images), which is measured by the Fr$\acute{e}$chet distance. Lower FID values suggest that the synthetic data distribution is much closer to the real data distribution. The lower FID, the higher realism of generated images. To compute the FID score, we randomly select 12,000 generated images for evaluation, i.e., for each test text, we randomly draw the noise vector four times and perform the generation. All images are resized to $64 \times 32$. (2) \textbf{Inception Score} (IS) \cite{Inception-score} measures the objectiveness and the diversity of generated images: IS=$\exp{(\mathbb{E}_x D_{KL} (p(y|\mathbf{x})||p(y)))}$, where $\mathbf{x}$ denotes one generated example, and $y$ is the label predicted by the inception model \cite{Going-deeper} \footnote{In our experiments, we directly use the pre-trained inception model \cite{Going-deeper}.}. To compute the IS, we follow the experimental setting of StackGAN-v2 \cite{StackGAN-pp} to sample 3,000 $256\times 128$ images. (3) \textbf{Visual-Semantic Similarity} (VS-Sim) \cite{HDGAN} measures the alignment between the generated images and the conditioning text, i.e., the semantic consistency. Following \cite{HDGAN}, we train a visual-semantic embedding model, which is used to measure the distance between synthesized images and the input text. More specifically, let $\mathbf{v}$ denote an image feature vector extracted from an Inception model $g(\cdot)$ \cite{Going-deeper}, we define a scoring function $s(\textbf{x},\textbf{y})=\frac{\textbf{x}\cdot \textbf{y}}{||\textbf{x}||_2 \cdot ||\textbf{y}||_2}$. Two mapping functions, i.e., $f_v(\cdot)$ and $f_c(\cdot)$ are trained to map the real images and paired text embedding into a common space with 512 dimensionality, and the following bi-directional ranking is loss is minimized:
\begin{equation}\label{eq:ranking-loss}
\begin{split}
& \mathcal{L}_{\mbox{rank}} = \sum_{\textbf{v}} \sum_{\bar{\textbf{c}}}\max [0,\delta-s(f_v(\textbf{v}), f_t(\textbf{t})) \\
& + s(f_v(\textbf{v}), f_c(\bar{\textbf{c}})) ]
    + \sum_{\textbf{c}} \sum_{\bar{\textbf{v}}}\max [0,\delta-s(f_c(\textbf{c}),f_v(\textbf{v})) \\
 &   + s(f_c(\textbf{c}), f_v(\bar{\textbf{v}})) ],
\end{split}
\end{equation}
where $\delta$ is the margin, which is set to 0.2. $\{\textbf{v}, \textbf{c}\}$ is a positive image-text pair, while $\{\textbf{v}, \bar{\textbf{c}}\}$ and $\{\bar{\textbf{v}}, \textbf{c}\}$ denote the mis-matched image-text pairs. In testing, given a text embedding $\textbf{c}$ and the generated image $\textbf{x}$, the VS score can be calculated as $s(g(\textbf{x}),\textbf{c})$. A higher VS score indicates better semantic consistency.


\begin{figure*}[t]
\includegraphics[width=2\columnwidth]{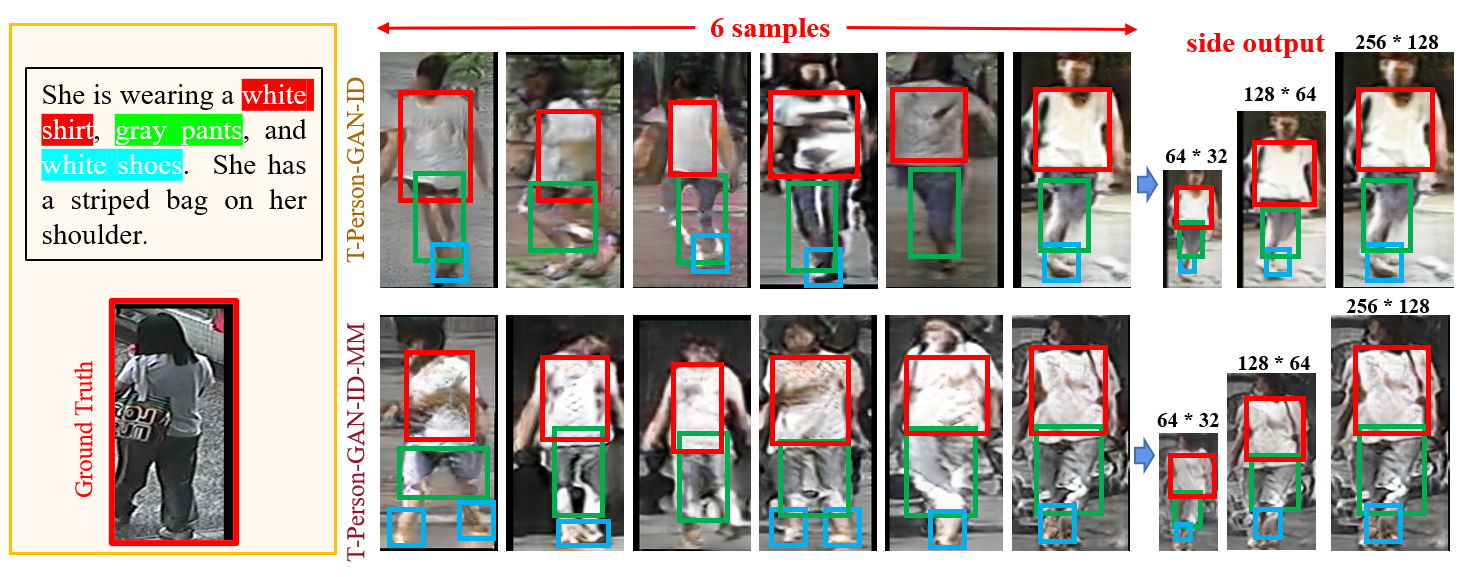}\\
\includegraphics[width=2\columnwidth]{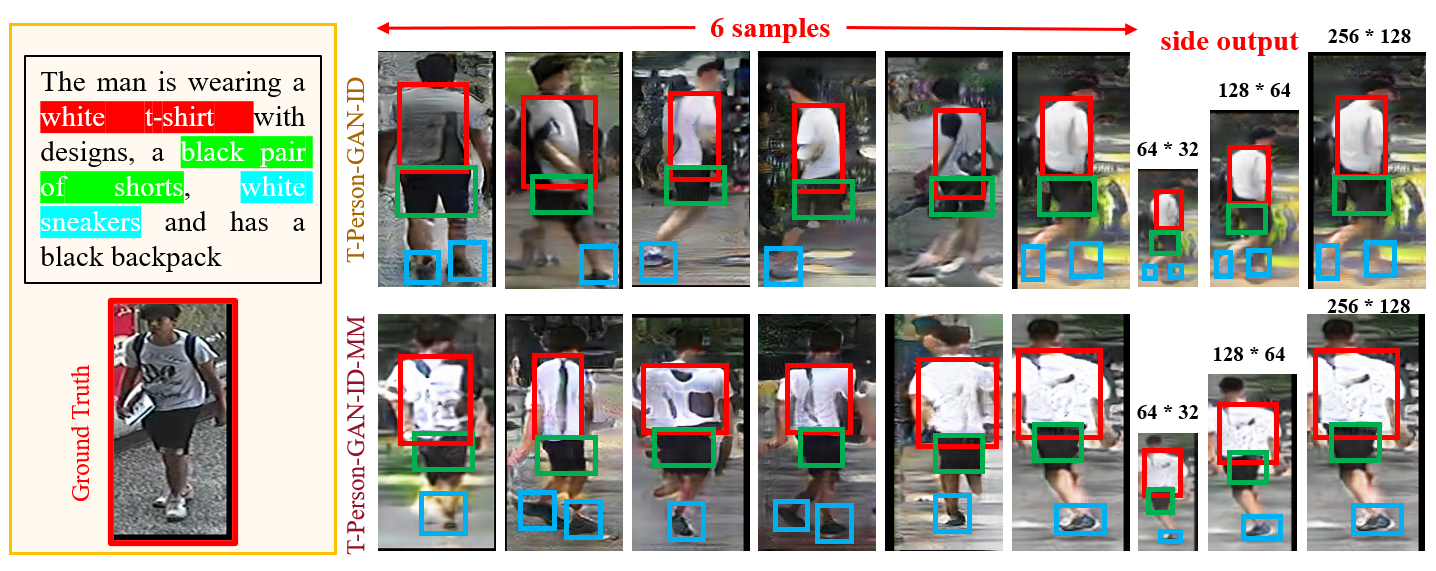}
\caption{The proposed methods (T-Person-GAN-ID and T-Person-GAN-ID-MM) can generate multiple samples given a single text input. T-Person-GAN-ID generates identity-preserving samples with better fine-grained details. T-Person-GAN-ID-MM advances the generation by synthesizing sharper details which are highly discriminant among identities. The side outputs of the proposed methods with increasing resolutions exhibit semantically consistent.}\label{fig:side-output}
\end{figure*}

\subsection{Comparison with Text-to-Image Generation Models}

As there is no existing methods that are exactly comparable with this work, we implement four popular baselines for text-to-image generation: GAWWN \cite{GAWWN}, GAN-INT-CLS \cite{GAN-CLS}, AttnGAN \cite{AttnGAN}, HDGAN \cite{HDGAN}, and StackGAN-v2 \cite{StackGAN-pp}. For all the baselines, we use the source codes provided by the authors, and the pre-trained models are available in Github \footnote{https://github.com/linwu-github/Person-Image-Generation.git}. We remark that StackGAN-v2* is re-implemented by combining StackGAN-v2 \cite{StackGAN-pp} with a normalization method known as spectral normalization \cite{Spectral-norm}, so as to stabilize the training of the multi-stream generative model. The spectral normalization \cite{Spectral-norm} only normalizes the weight matrices of discriminator, which is compatible to any generative model without impacting its task performance. We also consider three recent SOTA methods: MirrorGAN \cite{MirrorGAN}, RiFeGAN \cite{RiFeGAN}, DF-GAN \cite{DF-GAN} and SSA-GAN \cite{SSA-GAN}.

The comparison results with aforementioned generative methods are reported in Table \ref{tab:main-evaluations}. First, both GAWWN \cite{GAWWN} and GAN-INT-CLS \cite{GAN-CLS} are early text-to-image generation methods for center-object generation, whilst they are limited to generating high-resolution images. Comparing with StackGAN-v2* \cite{StackGAN-pp} and AttnGAN \cite{AttnGAN} that are based on a multi-steam architecture, the proposed approaches (T-Person-GAN-ID and T-Person-GAN-ID-MM) outperform these state-of-the-arts across all evaluation metrics. For instance, StackGAN-v2* \cite{StackGAN-pp} achieves FID=61.62 and IS=3.53. In contrast, the T-Person-GAN-ID achieves a lower FID and a higher IS than the former. This shows an obvious advantage of using one-stream generator. Moreover, in comparison with a strong baseline based on one-stream architecture, i.e., HDGAN \cite{HDGAN}, the proposed T-Person-GAN-ID reduces the FID value from 61.62 to 48.04. The proposed manifold mix-up mechanism (T-Person-GAN-ID-MM) achieves the lowest FID=47.81. Meanwhile, the IS is increased from 3.63 (HDGAN \cite{HDGAN}) to 3.92 (T-Person-GAN-ID-MM). Finally, in VS-Sim evaluation, AttnGAN \cite{AttnGAN} achieves superior performance to other methods, because AttnGAN \cite{AttnGAN} considers the textual-visual alignment during the generation and the objective function is derived with respect to such alignment. In contrast, our method deploys multiple discriminators, each of which can strengthen the visual details with semantically meaningful.

Comparing with more recent SOTA methods, i.e., MirrorGAN \cite{MirrorGAN}, RiFeGAN \cite{RiFeGAN}, DF-GAN \cite{DF-GAN} and SSA-GAN \cite{SSA-GAN}, our methods consistently achieves the superior performance across both FID and IS metrics. It is worth mentioning that all these methods need additional priors to improve the generation quality on the center-object generation. For example, SSA-GAN \cite{SSA-GAN} requires a weakly-supervised mask to guide where to generate the local regions. RiFeGAN \cite{RiFeGAN} firstly exploits an attention-based caption matching model to refine the captions from prior knowledge. And then the image generation is based on such enriched texts. Nonetheless, our proposed methods do not require any priors or additional processing on texts.

\begin{figure*}[h]
\centering
\includegraphics[width=11cm,height=8cm]{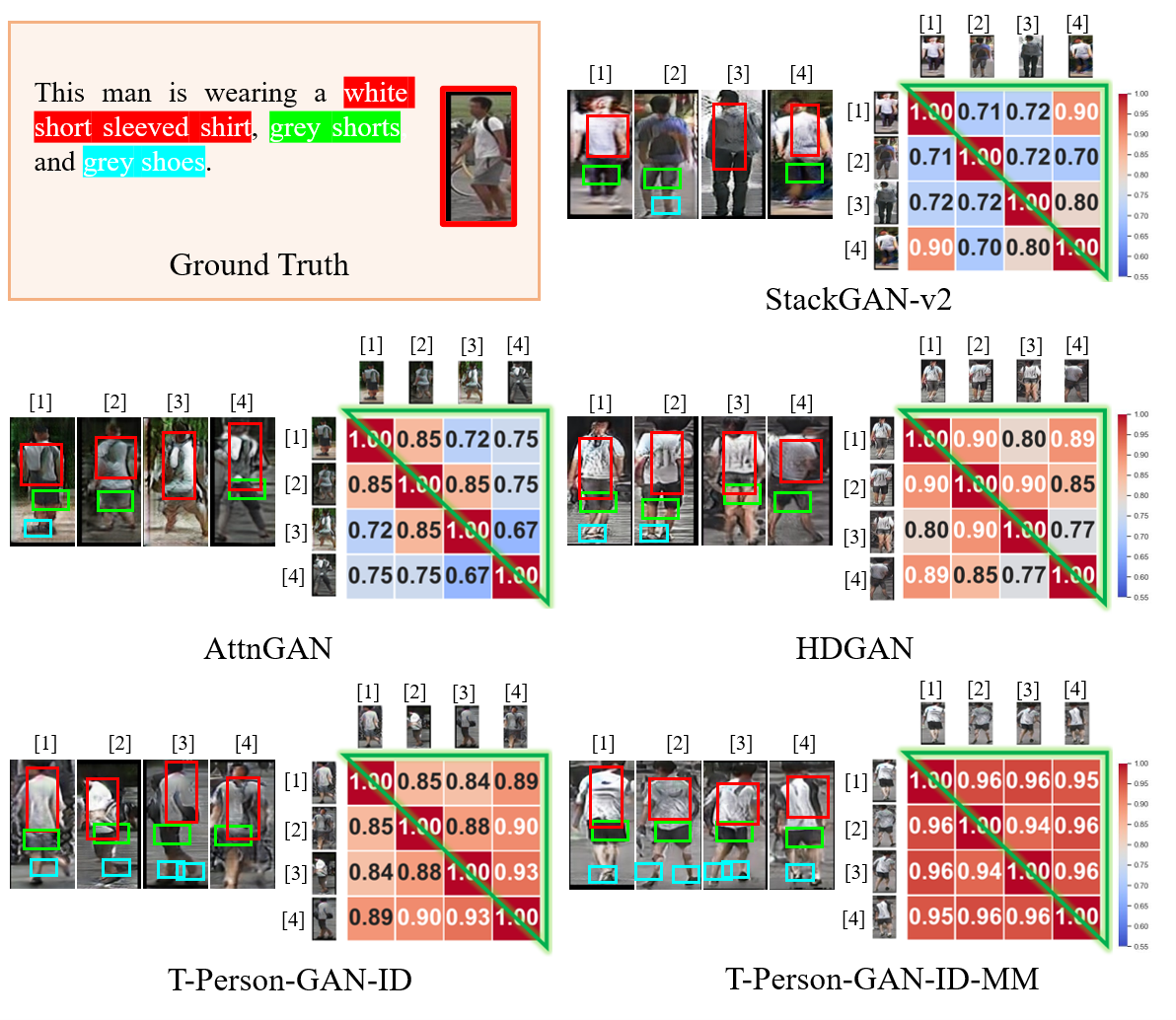}\\
\includegraphics[width=11cm,height=8cm]{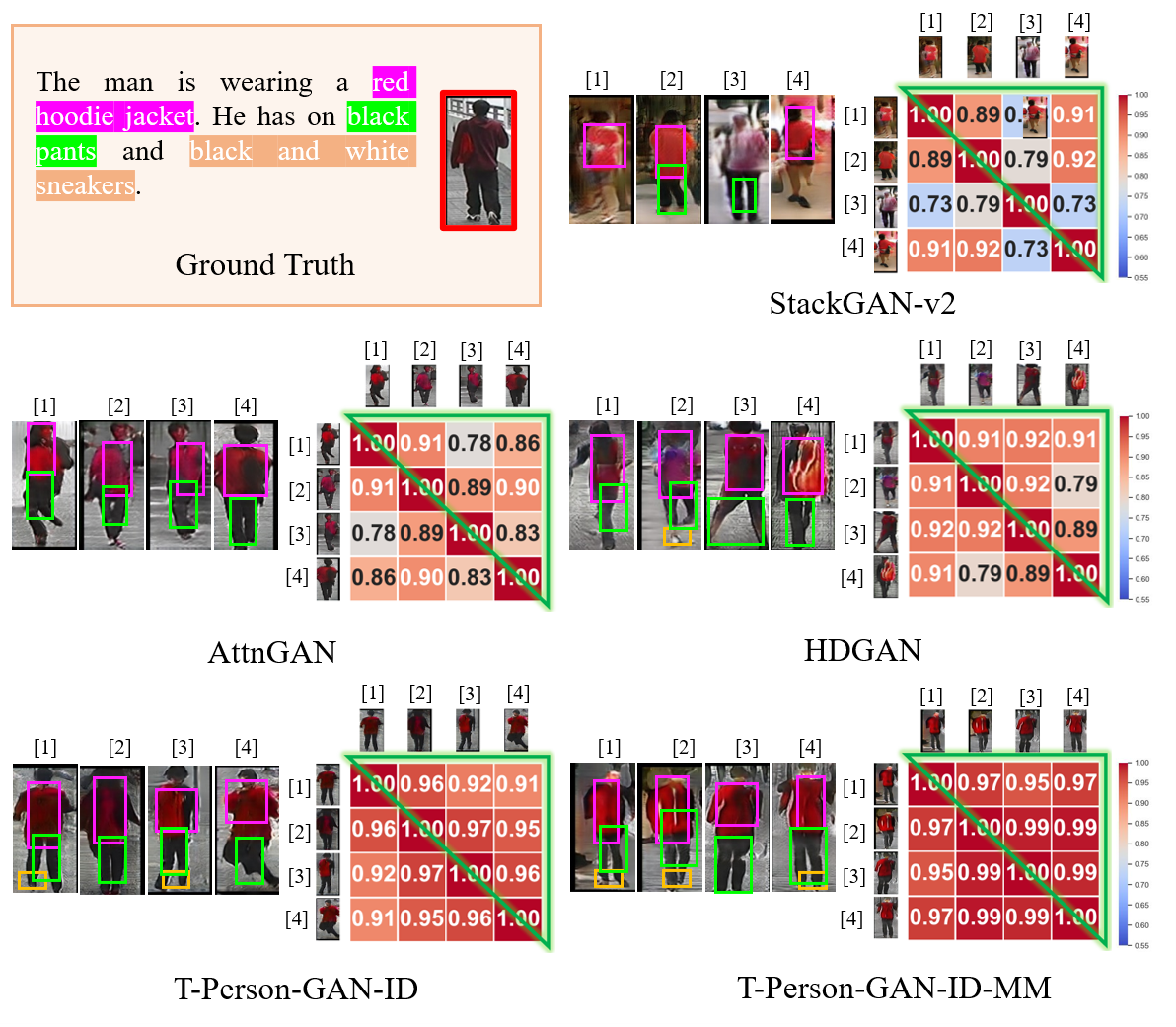}
\caption{Comparison results on the CUHK-PEDES dataset. Center-object generation methods (StackGAN-v2 \cite{StackGAN-pp}, AttnGAN \cite{AttnGAN} and HDGAN \cite{HDGAN}) use randomness to generate diverse images, resulting in identity inconsistent and less correlation amid the generated person images in feature space. Our methods (T-Person-GAN-ID and T-Person-GAN-ID-MM) synthesizes person image with identity-consistent and discriminant. These generated images are highly correlated in the feature space. }\label{fig:CUHK-generated}
\end{figure*}


The qualitative comparison results are shown in Fig. \ref{fig:CUHK-generated}, where a sequence of person images synthesized by different generative models are compared. We consider the following center-object baselines: StackGAN-v2 \cite{StackGAN-pp}, AttnGAN \cite{AttnGAN}, and HDGAN \cite{HDGAN}. We can make the following observations. First, center-object generation methods mainly adopt the randomness to produce diverse outcomes. For person image generation, these methods tend to generate inconsistent regions in person images. For example, in Fig. \ref{fig:CUHK-generated}, StackGAN-v2 \cite{StackGAN-pp} is unable to generate four images with each image containing all the important patterns, i.e., white short sleeved shirt, grey shorts/shoes. This result can also be observed in AttnGAN \cite{AttnGAN} which focuses on the visual-text alignment via attention. To verify this observation, we study the affinity matrix between the generate images. More specifically, we consider four generated person images, and for each image we use ResNet-50 \cite{Resnet} to extract the feature from the penultimate layer. To calculate the affinity value between two images, we use the cosine distance. For StackGAN-v2 \cite{StackGAN-pp} and AttnGAN \cite{AttnGAN}, the lower affinity values indicate that these generated samples are less correlated in feature space. The similar observation can also be made in HDGAN \cite{HDGAN}. Such a practice works for center-object generation like birds and dogs, but incapable of generating person images. In stark contrast, our methods can synthesize person images which are highly identity-consistent and yet discriminant against different identities under visual variations. These experimental results clearly verify our observation, as described in Section \ref{ssec:ID-network} and \ref{ssec:manifold-mix-up}.

\begin{figure*}[hbt]
\begin{tabular}{ccc}
\includegraphics[width=6cm,height=5cm]{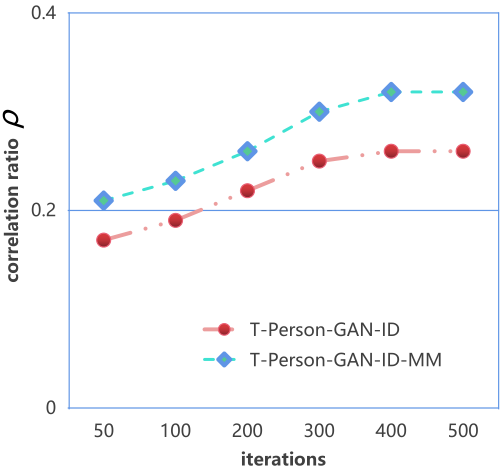}
     & \includegraphics[width=6cm,height=5cm]{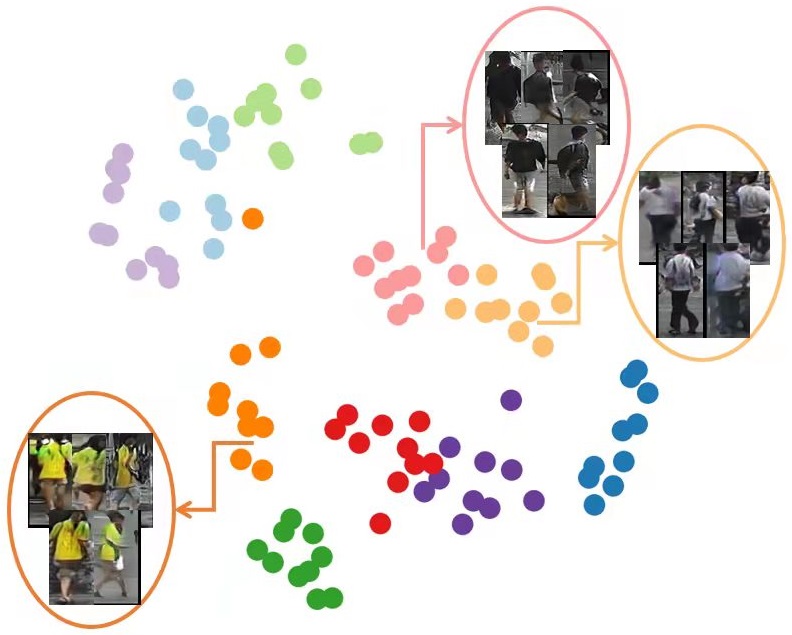}
    & \includegraphics[width=6cm,height=5cm]{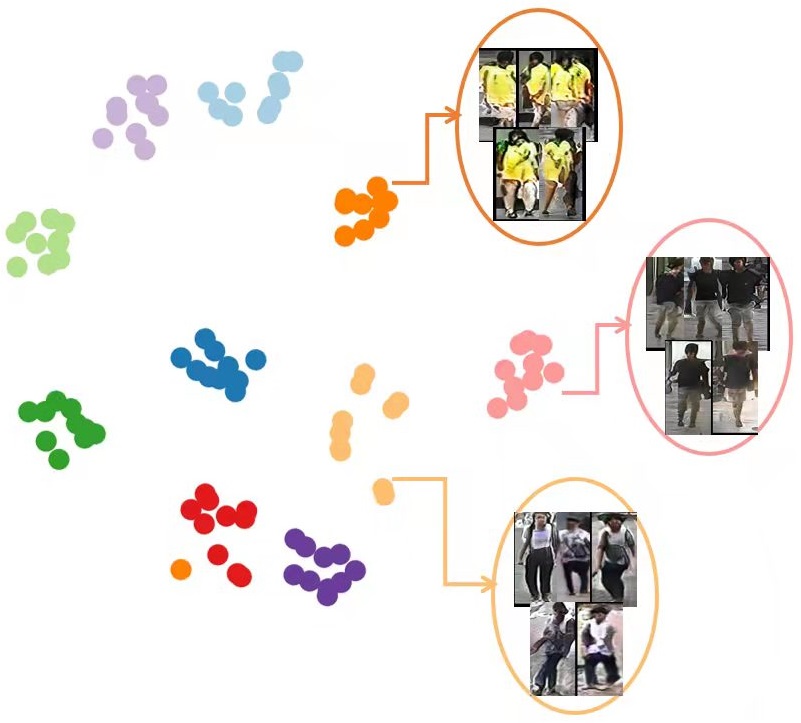}\\
    (a) & (b) & (c)
\end{tabular}
\caption{(a) The correlation ratio computed over the generated data vector matrix of two methods. (b) t-SNE embedding of T-Person-GAN-ID. (c) t-SNE embedding of T-Person-GAN-ID-MM. The correlation ratio increases as more iterations running. In the embedding space, both T-Person-GAN-ID and T-Person-GAN-ID-MM ensure the strong correlation amid the generated images within one identity. It is remarked that T-Person-GAN-ID-MM shows even stronger correlation than T-Person-GAN-ID. Best viewed in color.}\label{fig:correlation-t-SNE}
\end{figure*}

\begin{figure}[t]
\begin{tabular}{cc}
\hspace{-1cm}\includegraphics[width=5cm,height=4cm]{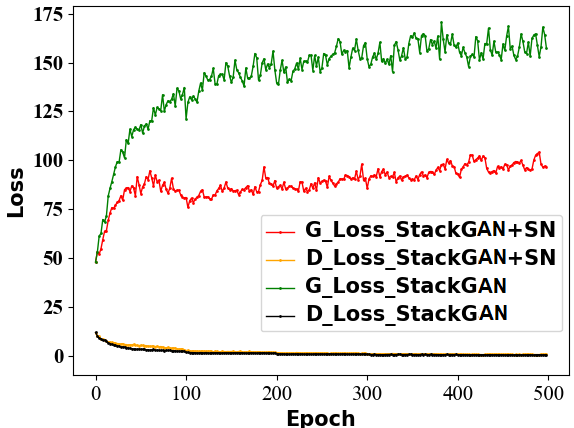}
& \hspace{-0.3cm}\includegraphics[width=5cm,height=4cm]{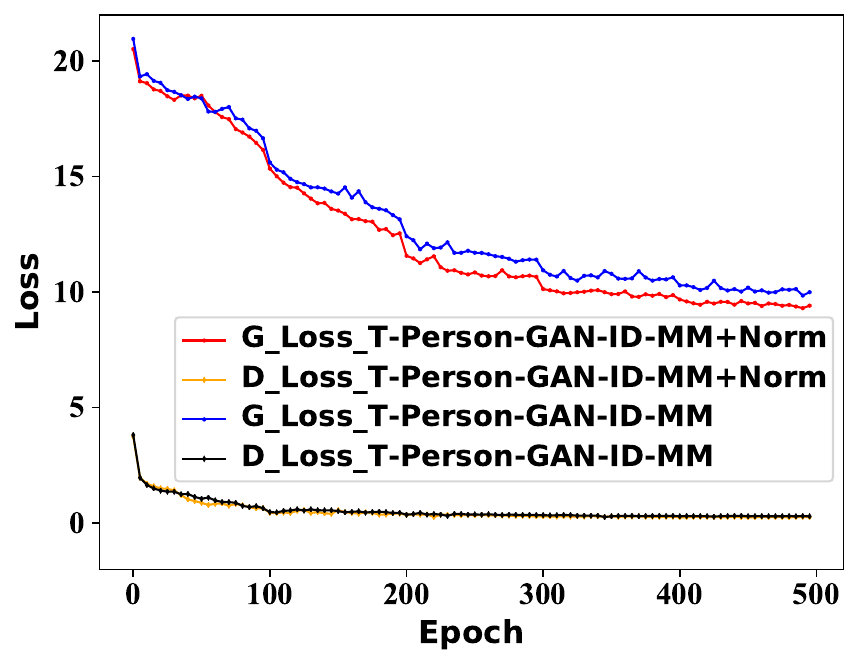}
\end{tabular}
\caption{Multi-stream generation vs. one-stream generation with/without normalization. Left: StackGAN-v2 (multi-stream) with/without spectral normalization on weights. Right: The proposed T-Person-GAN-ID-MM (one-stream) with/without input normalization.}\label{fig:Loss-with-norm}
\end{figure}

\subsection{Why One-Stream Architecture is Good for Person Image Generation?}

One may wonder the effectiveness of one-stream architecture in producing the resolution-increasing person images. To verify this, we show the advantage of adopting such an architecture, which is able to achieve identity consistency pertaining to person images. As shown in Fig. \ref{fig:side-output}, it demonstrates the consistency of our method in generating visual outputs with high identity relevance. Following on, we show a series of images at increasing resolutions as the side output of the one-stream generator. For instance, given the input texts ``She is wearing a white shirt, gray pants, and white shoes. She has a stripped bag on her shoulder", our approaches present the figures correctly in its pattern and colour accordingly. It shows that our approaches are able to produce fine-grained visual details reasonably corresponding to the given texts. This also highlights the benefit of identity-preserving component to allow for visual-semantic generation across resolutions.

\subsection{Ablation Studies}

\subsubsection{Manifold Mix-up for Linear Separation in Embedding}

In this experiment, we study the correlation between the generated data in the feature space. With this purpose, the proposed ID-preserving mechanism (T-Person-GAN-ID) and the improved manifold mix-upping (T-Person-GAN-ID-MM) are evaluated both quantitatively and qualitatively. More specifically, we randomly choose 10 texts corresponding to 10 different identities from the testing set. Given each text as input, the image generation is performed 100 times by randomly drawing 100 Gaussian noises. Then, the noise vector is concatenated with the text embedding to feed the generator. The correlation ratio $\rho$ is calculated using the data vector matrix formed by the generated images for each identity. As shown in Fig. \ref{fig:correlation-t-SNE} (b) and (c), the t-SNE embeddings of both mechanisms are well-separated for each identity. This linear separation in embedding is more obvious for T-Person-GAN-ID-MM, which shows clear linear boundary between different identities. Quantitatively, the correlation ratio for the two mechanisms (Fig. \ref{fig:correlation-t-SNE} (a)) exhibits the strong correlation amid these generated images within each identity. This experiment verifies that T-Person-GAN-ID-MM achieves a better separation between different identities. This is boiled down to the effectiveness of linear interpolation, which suggests robustness against inter-personal variations.

\subsubsection{One-Stream Generator vs. Multi-Stream Generators: How To Stabilize Person Image Generation?}

Generating high-resolution person images using texts only has to deal with \textit{complex} pattern statistics. StackGAN-v2 \cite{StackGAN-pp} provides a viable solution to the central-object generation (e.g., flowers and birds) by associating each resolution with its corresponding generator (coupled with a discriminator). Such a multi-stream approach is less effective in stabilizing the GAN due to the difficulty of synchronizing the gradients from different streams. In the training of GANs, a generator needs the gradients provided by the discriminator. If the discriminator can perfectly discriminate two distributions at early steps of training, the generator would stop its learning as the saturation occurs. However, a multi-stream backbone such as StackGAN-v2 \cite{StackGAN-pp} cannot synchronize multiple discriminators. To validate the necessity of one-stream generator, we examine the training behaviours of two baselines: StackGAN-v2 \cite{StackGAN-pp}, and StackGAN-v2 \cite{StackGAN-pp} combined with spectral normalization on weights \cite{Spectral-norm}, i.e., StackGAN-v2+SN. We also examine the training of our method T-Person-GAN-ID-MM and with the input normalization on discriminator. The comparison results are shown in Fig. \ref{fig:Loss-with-norm}. The following observations can be made. Firstly, when the StackGAN-v2 \cite{StackGAN-pp} is regularized by the spectral normalization on the weights of multiple discriminators, the generator loss is constrained rather than tending to be exploded if without spectral normalization. Secondly, the generator's loss of the proposed method shows a constant convergence when the input normalization is applied into the discriminator optimization. This helps searching a better discriminator in the min-max game.

\subsubsection{The Study on Manifold Mix-up Parameter}

\begin{table}[t]\scriptsize
\begin{minipage}[b]{0.56\linewidth}
\begin{tabular}{|c|c|c|}
\hline
Para & Epochs & Acc \%\\
\hline
\multirow{3}{*}{$\alpha$=0.1} &  100 & 2.56\\
& 300 & 22.01\\
& 500 & 22.72 \\
\hline
\multirow{3}{*}{$\alpha$=0.2} &  100 & 1.15\\
& 300 & 14.37\\
& 500 & 16.21 \\
\hline
\multirow{3}{*}{$\alpha$=0.4} &  100 & 1.20\\
& 300 & 6.97\\
& 500 & 11.2 \\
\hline
\multirow{3}{*}{$\alpha$=0.5} &  100 & 1.02\\
& 300 & 5.72\\
& 500 & 6.71 \\
\hline
\end{tabular}
\caption{Validation on $\alpha$}\label{table:alpha-acc}
\end{minipage}\hfill
\begin{minipage}[b]{0.4\linewidth}
\hspace{-1.5cm}\includegraphics[width=50mm]{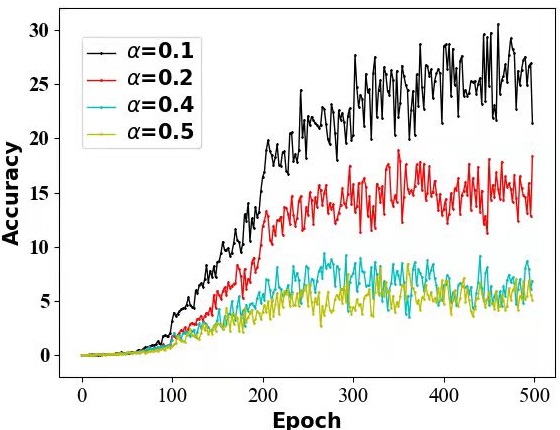}
\captionof{figure}{$\alpha$ w.r.t accuracy.}
\label{fig:alpha-acc}
\end{minipage}
\end{table}


The mix-up parameter $\lambda$ in Eq. \eqref{eq:mixup} is drawn from the $Beta(\alpha, \alpha)$ distribution, which is determined by $\alpha$. To study the effect of $\alpha$ that manipulates the interpolation extent between two identity generated samples, we conduct the experiment by evaluating the prediction error on mix-upped samples. Specifically, we randomly choose 1,000 training identities, for each of which we generate 54 samples at different resolution. Then, these generated data at each resolution are interpolated by a specific $\alpha \in [0.1,0.2,0.4]$. We train a ResNet-50 with 1,000-way classification to predict the label of each interpolated sample. The validation results are provided in Table \ref{table:alpha-acc}. We can see that when the mix-up component is trained for 200 epochs with $\alpha=0.1$, the accuracy of the mix-up scheme can be improved by 1.0 \% (3.5\%) compared to the 200 epochs ran by $\alpha=0.2$ ($\alpha$=0.3). However, a higher accuracy indicates a less interpolation between generated images since the classifiers can easily predict the correct categories for non-interpolated samples. Thus, to balance the accuracy and interpolation effect, we set $\alpha=0.2$ in all of our experiments, if not specified.

\section{Conclusion}\label{sec:con}
In this paper, we present a principled approach to generate discriminant person images from texts only. The method is based on an one-stream generative architecture, which integrates hierarchically-nested discriminators on respective resolution. To achieve high-quality person images, we propose two effective mechanisms: identity-preserving network and manifold mix-up interpolation. The identity-preserving mechanism promote the generated images with respect to identity characteristics. The manifold mix-up performs linear interpolation between generated data so as to separate the class boundary between identities. Extensive experiments on person image synthesis demonstrate our method is superior to state-of-the-arts in generating more realistic yet discriminant person images. We note that our architecture is orthogonal to StackGAN++ \cite{StackGAN-pp}, and focuses on person image generation, with all of them together to enrich the system of GANs for the image generation task.

%

\section*{Acknowledgment}
The authors would like to thank the constructive comments from both the review team and the editor. This work was partially supported by NSFC U19A2073, 62002096.

\ifCLASSOPTIONcaptionsoff
  \newpage
\fi



\bibliographystyle{IEEEtran}
\bibliography{allbib}

\end{document}